\documentclass[letterpaper, 10 pt, conference]{ieeeconf}
\IEEEoverridecommandlockouts                              
\usepackage{comment}
\usepackage{url}

\usepackage{multirow}
\usepackage{graphicx}
\usepackage{epsfig}
\usepackage{epic,eepic}
\usepackage{times}
\usepackage{mathptmx}
\usepackage{cite}
\usepackage{color}

\usepackage{times}

\definecolor{red}{rgb}{1,0,0}
\definecolor{green}{rgb}{0,1,0}
\definecolor{blue}{rgb}{0,0,1}
\definecolor{violet}{rgb}{1,0,1}
\definecolor{cyan}{cmyk}{1,0,0,0}
\definecolor{magenta}{cmyk}{0,1,0,0}
\definecolor{yellow}{cmyk}{0,0,1,0}

\definecolor{white}{rgb}{1,1,1}

\usepackage{arydshln}

\newcommand{\CO}[1]{}

\newcommand{\CommentOut}[1]{}

 \newcommand{\editage}[1]{}

\newcommand{\islarge}{}

\begin{document}


\newcommand{\FIG}[3]{
\begin{minipage}[b]{#1cm}
\begin{center}
\includegraphics[width=#1cm]{#2}\\
{\scriptsize #3}
\end{center}
\end{minipage}
}

\newcommand{\FIGU}[3]{
\begin{minipage}[b]{#1cm}
\begin{center}
\includegraphics[width=#1cm,angle=180]{#2}\\
{\scriptsize #3}
\end{center}
\end{minipage}
}

\newcommand{\FIGm}[3]{
\begin{minipage}[b]{#1cm}
\begin{center}
\includegraphics[width=#1cm]{#2}\\
{\scriptsize #3}
\end{center}
\end{minipage}
}

\newcommand{\FIGR}[3]{
\begin{minipage}[b]{#1cm}
\begin{center}
\includegraphics[angle=-90,width=#1cm]{#2}
\\
{\scriptsize #3}
\vspace*{1mm}
\end{center}
\end{minipage}
}

\newcommand{\FIGRpng}[5]{
\begin{minipage}[b]{#1cm}
\begin{center}
\includegraphics[bb=0 0 #4 #5, angle=-90,clip,width=#1cm]{#2}\vspace*{1mm}
\\
{\scriptsize #3}
\vspace*{1mm}
\end{center}
\end{minipage}
}

\newcommand{\FIGCpng}[5]{
\begin{minipage}[b]{#1cm}
\begin{center}
\includegraphics[bb=0 0 #4 #5, angle=90,clip,width=#1cm]{#2}\vspace*{1mm}
\\
{\scriptsize #3}
\vspace*{1mm}
\end{center}
\end{minipage}
}

\newcommand{\FIGpng}[5]{
\begin{minipage}[b]{#1cm}
\begin{center}
\includegraphics[bb=0 0 #4 #5, clip, width=#1cm]{#2}\vspace*{-1mm}\\
{\scriptsize #3}
\vspace*{1mm}
\end{center}
\end{minipage}
}

\newcommand{\FIGtpng}[5]{
\begin{minipage}[t]{#1cm}
\begin{center}
\includegraphics[bb=0 0 #4 #5, clip,width=#1cm]{#2}\vspace*{1mm}
\\
{\scriptsize #3}
\vspace*{1mm}
\end{center}
\end{minipage}
}

\newcommand{\FIGRt}[3]{
\begin{minipage}[t]{#1cm}
\begin{center}
\includegraphics[angle=-90,clip,width=#1cm]{#2}\vspace*{1mm}
\\
{\scriptsize #3}
\vspace*{1mm}
\end{center}
\end{minipage}
}

\newcommand{\FIGRm}[3]{
\begin{minipage}[b]{#1cm}
\begin{center}
\includegraphics[angle=-90,clip,width=#1cm]{#2}\vspace*{0mm}
\\
{\scriptsize #3}
\vspace*{1mm}
\end{center}
\end{minipage}
}

\newcommand{\FIGC}[5]{
\begin{minipage}[b]{#1cm}
\begin{center}
\includegraphics[width=#2cm,height=#3cm]{#4}~$\Longrightarrow$\vspace*{0mm}
\\
{\scriptsize #5}
\vspace*{8mm}
\end{center}
\end{minipage}
}

\newcommand{\FIGf}[3]{
\begin{minipage}[b]{#1cm}
\begin{center}
\fbox{\includegraphics[width=#1cm]{#2}}\vspace*{0.5mm}\\
{\scriptsize #3}
\end{center}
\end{minipage}
}









\islarge{
\LARGE
}

\title{\LARGE \bf
Recursive Distillation for Open-Set Distributed Robot Localization
}

\author{%
Kenta Tsukahara ~~~~~ Kanji Tanaka ~~~~%
}

\maketitle

\newcommand{\figA}{
\begin{figure}
\begin{center}
\FIGR{7}{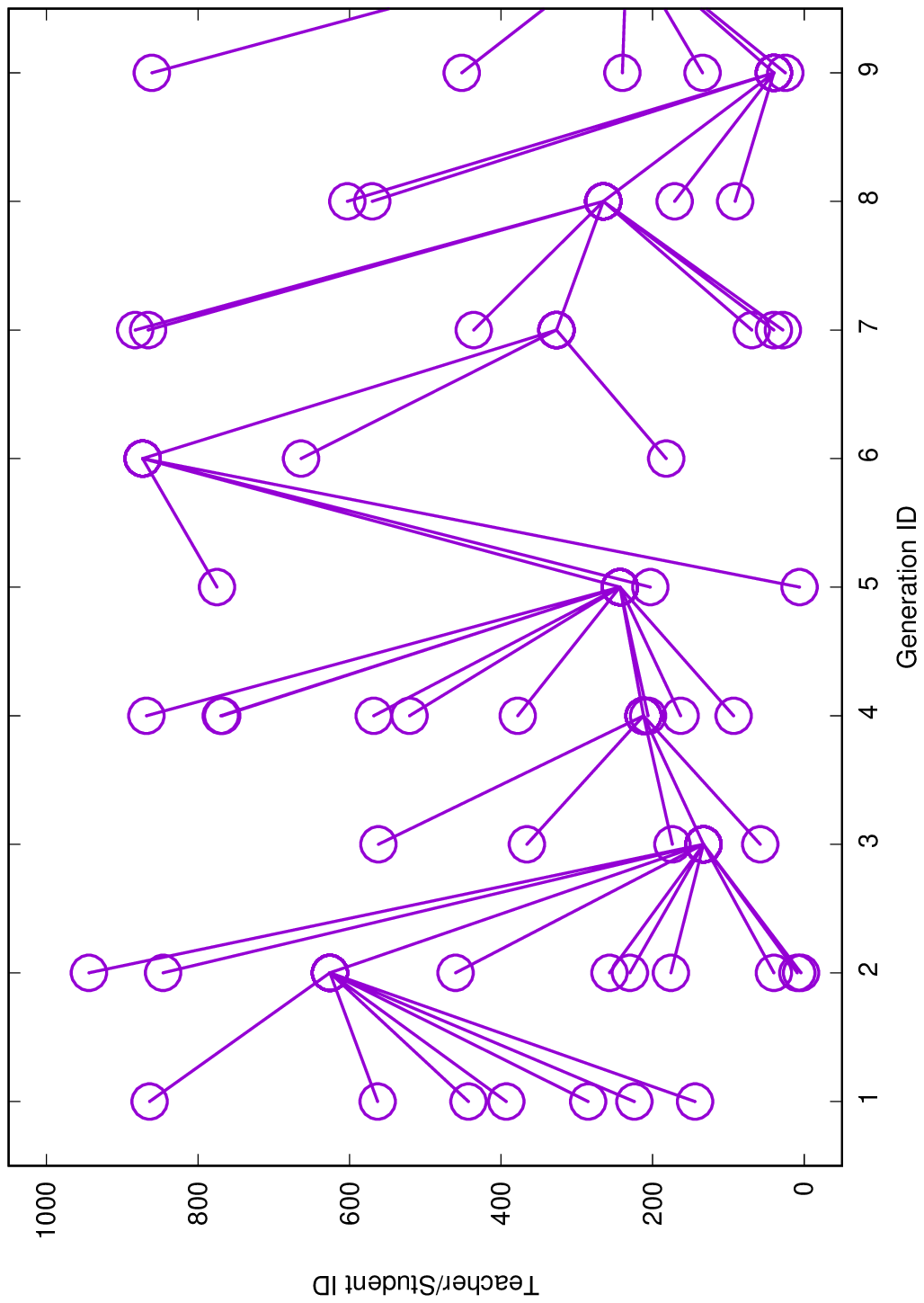}{}\vspace*{-5mm}\\
\caption{Recursive distillation scheme. Teacher/student models and teacher-to-student relationships are represented by circles and line segments. A student can ask an open teacher set to transfer knowledge. A trained student can recursively join the next-generation open teacher set.
}\label{fig:A}\vspace*{-8mm}~\\
\end{center}
\end{figure}
}

\newcommand{\figB}{
\begin{figure}
\begin{center}
\FIG{8}{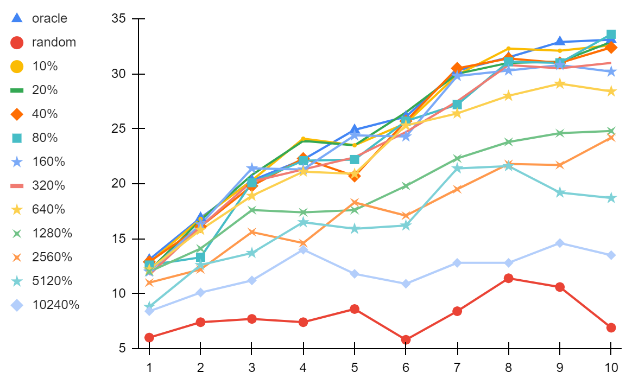}{}
\caption{Performance results. Self-localization performance of the student model at each generation is evaluated. The vertical and horizontal axes are Top-1 accuracy performance and generation ID, respectively.}\label{fig:B}
\end{center}
\end{figure}
}

\newcommand{\figC}{
\begin{figure}
\begin{center}
\FIG{4}{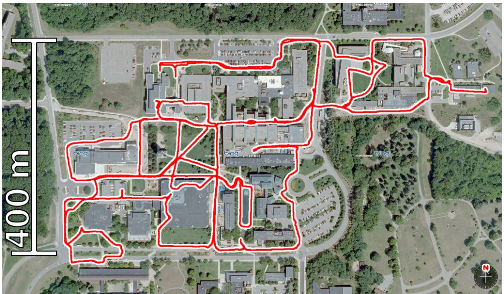}{}\FIG{4}{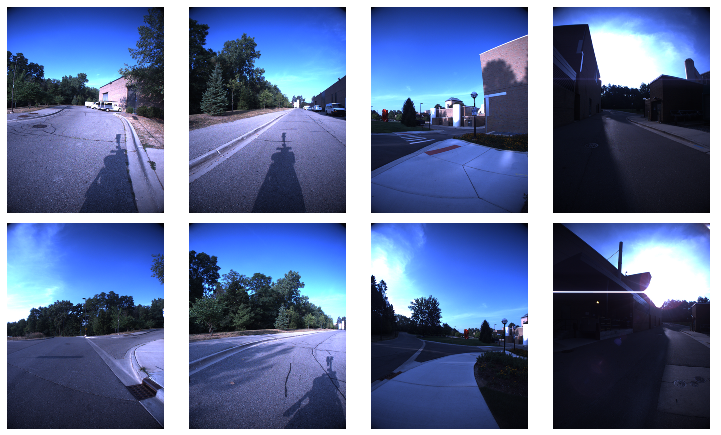}{}
\caption{NCLT dataset. 
Top: A bird's eye view of the robot workspace. 
Bottom: Sample view images from the robot's onboard camera.}\label{fig:C}~\vspace*{-5mm}
\end{center}
\end{figure}
}

\newcommand{\figD}{
\begin{figure}
\begin{center}
\FIG{8}{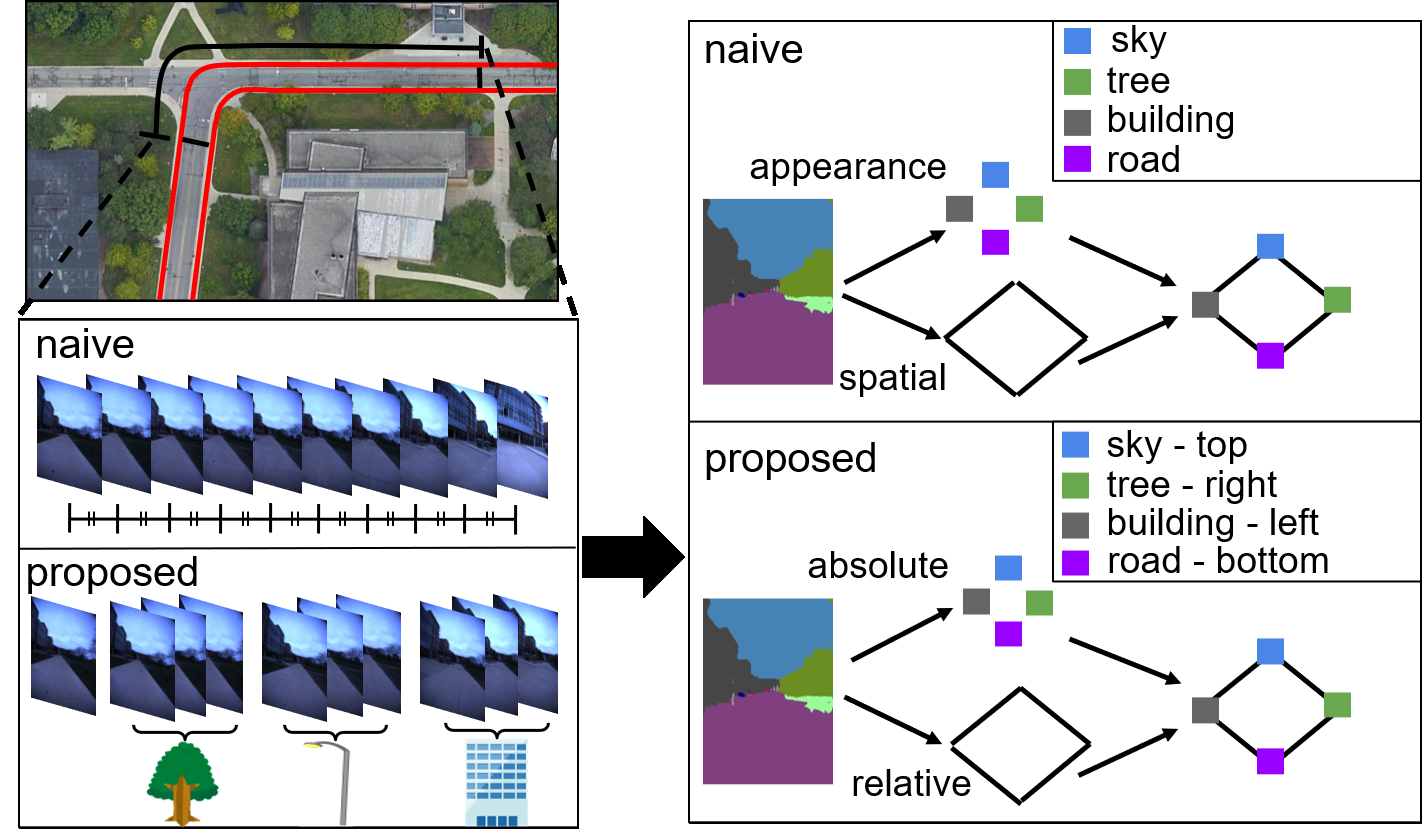}{}~\vspace*{-3mm}\\
\caption{Visual embedding used in the experimental setup. A spatial-semantic scene graph is generated from the view image. For details, please refer to \cite{irosw2022yoshida}.}\label{fig:D}
\end{center}
\end{figure}
}

\CO{
} 

\begin{abstract}
A typical assumption in state-of-the-art self-localization models is that an annotated training dataset is available for the target workspace. However, this is not necessarily true when a robot travels around the general open world. This work introduces a novel training scheme for open-world distributed robot systems. In our scheme, a robot (``student") can ask the other robots it meets at unfamiliar places (``teachers") for guidance. Specifically, a pseudo-training dataset is reconstructed from the teacher model and then used for continual learning of the student model under domain, class, and vocabulary incremental setup. Unlike typical knowledge transfer schemes, our scheme introduces only minimal assumptions on the teacher model, so that it can handle various types of open-set teachers, including those uncooperative, untrainable (e.g., image retrieval engines), or black-box teachers (i.e., data privacy). In this paper, we investigate a ranking function as an instance of such generic models, using a challenging data-free recursive distillation scenario, where a student once trained can recursively join the next-generation open teacher set.
\end{abstract}

\section{Introduction}

Self-localization, i.e., the problem of classifying a view image into predefined classes, is a fundamental problem in visual robot navigation and has important applications including scene understanding, map building, and path planning. Most of the existing solutions, ranging from image retrieval engines \cite{ibowlcd} to ConvNet image classifiers \cite{planet}, aim to build a high-quality self-localization model using annotated training datasets as supervision. Many state-of-the-art techniques can achieve very good performance in such supervised settings. However, this is not the case for an unfamiliar workspace where no supervision is available. Thus, the problem is largely unsolved.

In this work, teacher-to-student knowledge transfer in general open-world distributed robot systems is considered as an alternative training setup. We observe that when humans travel around the open world, they often ask the people they meet in unfamiliar places for guidance. Therefore, we propose a similar knowledge transfer scheme, in which a student robot can view other robots encountered in unfamiliar places as potential teachers, and ask them to transfer knowledge about the places. It is noteworthy that there may exist various types of potential teacher robots. Some of them may be cooperative, but some others may not be. Some of them may be trainable (e.g., differentiable neural networks \cite{planet}), but some others may not be (e.g., image retrieval engines \cite{ibowlcd}). Some of them may have a known architecture, but some may have a black box architecture (i.e., data privacy). Therefore, we propose to introduce only minimal assumptions on the potential teacher robots.

Existing knowledge transfer frameworks typically relied on prior knowledge about the teacher model, such as training datasets and metadata \cite{KTsurvey}. For example, the multi-teacher multi-student knowledge transfer scheme in \cite{itsc2018fang} succeeded in training students to perform at least as well as the teacher by transferring the teacher's training data to the students. However, such schemes can potentially increase the cost of maintaining training data. A similar problem is being studied in the machine learning community as a more general issue called continual learning \cite{CL}. Many of existing approaches fall into the categories of ``regularization \cite{RegularizationCL}," ``replay \cite{ReplayCL}," and ``dynamic architecture \cite{DynamicArchitectureCL}," all of which require maintenance of training datasets and metadata. Such requirements significantly limit the range of applications.

\figA

To address the issue, we explore a novel continual learning setup that can handle an open teacher set. Instead of annotated training dataset to be required, in our scheme, a pseudo training dataset is reconstructed from a teacher model and used for continual learning of the student model under domain, class, and vocabulary incremental setups. However, even with current state-of-the-art technology \cite{DataFreeKnowledgeTransferSurvey}, dataset reconstruction abilities are far from perfect \cite{DatasetReconstruction}. Furthermore, teacher robots can be of various types, ranging from trainable models such as ConvNet image classifiers \cite{planet} to untrainable models such as view image retrieval engines \cite{ibowlcd}. Unfortunately, existing schemes assume a known supervised architecture (e.g., \cite{DataFreeKnowledgeTransferSurvey31}) and cannot be applied to open-set distributed robot systems. Therefore, we wish to find ``generic" teacher models that can handle not only known teachers but also untrainable teachers (e.g., image retrieval engines \cite{ibowlcd}) with unknown architectures (i.e., data privacy \cite{PrivacyDataFreeKT}). In this work, we present a ranking function as an instance of such generic teacher models and investigate its performance in a challenging data-free recursive distillation scenario \cite{itsc2019hiroki} (Fig. \ref{fig:A}), where a trained student can recursively join the next-generation open teacher set.

\figC

\section{Open-Set Distributed Robot Localization}\label{sec:problem}

The current approach is built on the multi-teacher multi-student knowledge transfer (KT) scheme in \cite{itsc2018fang}. In \cite{itsc2018fang}, an ensemble KT scheme was investigated for visual place classification. The NCLT dataset in \cite{nclt} (Fig. \ref{fig:C}) was used, which contains long-term navigation data of a Segway robot equipped with an onboard monocular front-facing camera navigating a university campus over a long period in over 20 seasonal domains. Figure \ref{fig:C} shows a bird's eye view of the robot workspace and example view images from the robot's onboard front-facing camera. The place classes are defined by partitioning the workspace into a grid in the bird's eye view coordinate system and associating each grid cell with each place class. Then, self-localization is formulated as a task of classifying an input view image to a place class. It is assumed that a teacher model has been trained in a previous season (i.e., domain). Then, the goal of KT is to train a student model so that it can be adapted to a new season via KT from the teachers.

Following \cite{itsc2018fang}, the knowledge to be transferred is assumed to be in the form of a training set (e.g., annotated visual inputs). This training set should be of the highest possible quality so that a model equivalent to the teacher model can be reconstructed from it. In the multi-teacher multi-student recursive KT scenario  \cite{itsc2018fang}, student robots can encounter multiple teachers sequentially. Once trained, students can also act as teachers for other potential students in subsequent seasons. Every time a student encounters a teacher, the student is provided with a training set by the teacher. In the original study, the training set is used for supervised learning of the student, assuming a ConvNet as the backbone of teacher/student models and a sample-based place class description \cite{iros2015kanji}. A variant with a knowledge distillation instead of supervised learning called ``recursive knowledge distillation" is also studied in \cite{itsc2019hiroki}. 

Unlike \cite{itsc2018fang}, the current work considers a data-free knowledge transfer (DFKT) extension of the recursive distillation scheme, called data-free recursive distillation (DFRD), to handle a general open teacher set that can contain unknown and black-box teachers. Recently, the research on DFKT has become very active in several research fields such as privacy-friendly KT \cite{PrivacyDataFreeKT}. In DFKT, training sets and metadata are not assumed to be accessible, but they should be reconstructed from the available teacher model and then used as pseudo-supervision for training the student model. The DFKT approach has a significant advantage in terms of spatial costs, as it does not require additional datasets or metadata like existing KT frameworks. This property is also attractive for the open-set distributed robot localization considered in the current study. In addition, DFKT also allows for the respect of the privacy of teacher robots and student robots. Specifically, students can hide what they don't know by minimizing their questions, and teachers can hide what they do know by minimizing their answers. Note that students do not necessarily have access to meta-information such as the number of teachers, the performance of individual teachers, the number and ID of unseen place classes, the relative pose of teachers, and physical means of communication.

\section{Data-Free Recursive Distillation}\label{sec:DFRD}

Our framework, data-free recursive distillation (DFRD), can be viewed as a DFKT extension of the recursive knowledge distillation, which was originally introduced in \cite{itsc2019hiroki}. However, unlike existing DFKT schemes, it can handle not only known teacher models but also generic teachers including untrainable and black-box teachers. Moreover, it allows the student to recursively act as a member of the next-generation open teacher set once trained. To our knowledge, such a data-free recursive distillation setup has not been considered in the existing works.

The basic idea of DFKT is to synthesize alternative data for the training data of the teacher model. Now, let $f$ be a teacher classifier that returns a prediction $y=f(x)$ for an input $x$.
In this case, the goal of DFKT is to reconstruct a high-quality training sample set $D=\{(x,y)\}$ from which a model equivalent to $f$ can be trained in a supervised manner using the pre-trained model $f$ as a cue.
In our experimental scenario, the data synthesizer $g$ is a trainable function that reconstructs the dataset $D$ such that $D=g(f)$. From the KT perspective, $D$ can be directly used for supervised training or distillation of the student model $f'$. The predictive performance of this trained student $f'$ depends on the quality of the synthetic data $D$. Under the assumption of ideal quality synthetic data $D$, students are expected to have the same or better prediction performance than the teacher. However, even state-of-the-art data synthesizers are far from perfect and are subject to reconstruction errors, including false positives and false negatives. As a result, the student's quality may be worse than the teacher's.

Existing frameworks on DFKT typically rely on individual assumptions on the teacher model $f$ (e.g., model type, architecture). For example, the seminal work in ZSKD \cite{DataFreeKnowledgeTransferSurvey31} introduces the hypothesis that softmax spaces can be modeled as Dirichlet distributions, and provides data impressions from the supervised model that can be used as a replacement for the training set. In DAFL \cite{DataFreeKnowledgeTransferSurvey38}, activations and softmax predictions are assumed to be available and this assumption is exploited to introduce a general-purpose regularization function for activations and predictions. These regularizations are also introduced in follow-up studies. Unfortunately, such assumptions of trainable or known teacher models are often violated in open-set teachers.

Instead of assuming such specific teacher models, we make a minimal assumption about the model: ``The teacher's self-localization model can be reused as the communication channel for KT." More specifically, the KT proceeds in the following procedure. (1) A student generates or samples a question $x$ and sends it to a teacher. (2) The teacher with model $f$ computes an answer by $y=f(x)$ and returns it to the student. (3) The student obtains a pseudo training sample $(x, y)$. One of the best-known strategies to generate a sample $x$ at Step 1 is to predict pseudo-training samples of the teacher or impressed as mentioned above. However, the problem of sampling impressions is highly ill-posed and a topic of ongoing research.

In this work, we begin with the best possible and worst possible samplers, called oracle samplers and random samplers. The oracle sampler has an excellent ability to sample $x$ from the teacher's training set (i.e., $x\in{D}$). The random sampler is a naive sampler that samples an input sample $x$ randomly from the input space. Oracle samplers and random samplers can be considered the best and worst-performing practical samplers that produce meaningful input samples. In other words, any practical sampler would be expected to perform somewhere between these two opposing samplers. Based on this consideration, we simulate diverse samplers by mixing sample sets from these two samplers at various mixing ratios. In experiments, the mixing ratio 100:$r$ was changed to $r$=$10\cdot{2^i}$ $(i=0, 1, 2, \cdots, 10)$.

Specifically, we proposed to use the reciprocal rank feature (RRF) as a regularized input signal. This is because the above naive random sampler was experimentally found to have extremely poor performance. Whether used alone or in combination with the Oracle sampler, the performance of the naive random sampler was so poor that the entire framework broke down. The proposed RRF was originally introduced in \cite{icte2022ohta} as an input feature. Recalling that any self-localization model can be modeled as a ranking function, any teacher's output sample or student's input sample can be approximately represented as an RRF vector. This RRF vector is low-dimensional and is well approximated by an even lower-dimensional $k$-hot RRF ($k$=10). Note that this $k$-hot RRF can be computed efficiently by performing $k$ maximum operations on an $N$-dimensional noise vector. It has been experimentally shown that this regularized random sampler is superior to the aforementioned naive random sampler. For more information on this RRF, please refer to the paper \cite{icte2022ohta}.

It is worth mentioning that from the continual learning perspective \cite{CL}, this work applies to all domain-, class- and vocabulary-incremental setups.
We assume a typical domain incremental scenario, which we call cross-season self-localization, in which a Segway robot navigates an outdoor workspace on a university campus over a long period.
We also assume a class increment scenario where students may incrementally learn a place class that is unknown to them from the teacher.
We also assume a vocabulary increment scenario in which students incrementally encounter diverse and unknown teacher vocabulary sequences during their travels (i.e., open vocabulary).
In other words,
it is assumed that
the open teacher set
is updated
when and only when
domain, class, or vocabulary
is incrementally updated.
Note that typical continual learning solutions ``regularization \cite{RegularizationCL}," ``replay \cite{ReplayCL}," and ``dynamic architecture \cite{DynamicArchitectureCL}" rely on the availability of datasets and metadata, and their extension to DFKT is a topic of ongoing research.

It is also worth mentioning that, unlike the typical well-defined place class definitions such as country/region/postal code, there are no criteria for clearly defining place classes in robot-centric coordinate systems.
There are various definitions and standards based on space and appearance, and they are not necessarily unified among robots.
Inconsistency in definitions and standards may become a serious obstacle in transferring class-specific knowledge between different robots.
As a naive solution, for example, when partitioning the robot's workspace into a grid of place classes by grid-based spatial discretization, a representative point such as the center of gravity of a place area in the workspace may be used as the unified definition of place class. 
However, such a place area represented by the same class ID is not necessarily spatially consistent between robots and may suffer from spatial uncertainties.
Therefore, the problem of KT-friendly place class definition remains an open issue \cite{iv2020kanji}.

\section{Experiments}

We experimentally evaluated the proposed scheme in a sequential cross-season scenario, using a length 10 sequence of seasons, ``2012/03/31,'' ``2012/1/8,'' ``2012/2/5,'' ``2012/2/23,'' ``2012/4/5,'' ``2012/6/15,'' ``2012/8/20,'' ``2012/10/28,'' ``2012/11/17,'' and ``2012/12/1'' in the NCLT dataset.
The details of the experimental setup are as in Section \ref{sec:problem}.
While our scheme also allows teachers and students to have different definitions of place classes (Section \ref{sec:DFRD}), for simplicity, all the teacher and student robots in the current experiments use the same place class definition, in which the workspace is partitioned into 100 place classes with a $10\times{10}$ grid of place classes.

\figD

For the self-localization model, a scene graph classifier recently developed in \cite{icte2022ohta} is employed as the visual embedding (Fig. \ref{fig:D}). It is a three-step procedure. (1) First, a spatial-semantic scene graph is extracted from an input view image by using a scene graph generator as in \cite{irosw2022yoshida}. (2) Then, the scene graph is converted to a class-specific probability map by using a pre-trained graph ConvNet. (3) Then, the class-specific probability map is further converted to an RRF format. Thus, such a fixed-length RRF vector is used as a visual input to the teacher and student models.

A minimal KT scenario that contains both supervised learning and knowledge transfer is considered.
A student model is initialized at each $i$-th season and the student robot encounters two teacher robots.
One of the two teachers is trained via supervised learning.
This teacher experienced a subset of the 100 place classes and employed the corresponding portion of the $i$-th season annotated dataset as supervision.
We randomly determined whether a teacher experienced a certain place class with a probability of 10\%, meaning that the number of place classes experienced by this teacher will be 10 on average. The other teacher is introduced for knowledge transfer. This teacher is the previous version of the student model that has been trained in $(i-1)$-th season. Note that this second type of teacher model is not available in the first season (i.e., $i=1$).

Note that for classes that the teacher has never experienced before, no matter how good the training scheme is, one can only expect a very low correct answer rate (about 1\%).
According to the above definition of experienced place classes, there can be overlap in the classes assigned to robots.
For example, in one experiment shown in Fig. \ref{fig:D}, the total number of classes teachers experienced over time was 10, 18, 25, 31, 34, 41, 44, 46, 47, and 48 for each generation. In this way, the number of classes experienced increases monotonically with the number of generations, but it is not strictly proportional.

The self-localization models are implemented as follows. The same multi-layer perceptron (MLP) model was used for all the teacher/student models. Given an input view image, the visual embedding is computed by the above-mentioned graph ConvNet, converted to a 10-hot RRF vector, and then used as input to the MLP model. The output of an MLP model is a class-specific softmax vector. However, it is converted to a class-specific rank vector to simulate the black-box teacher model. In preliminary experiments, the rank vector can be approximated by a 10-hot RRF vector to be used for knowledge distillation with the standard distillation loss function. However, we here consider a more generic scenario, and the RRF vector is further converted to a 1-hot vector. Finally, the pairing of the input embedding and the 1-hot vectors is used as a training sample for KT.

\figB

Figure \ref{fig:B} shows the performance curve. As can be seen from this figure, when the ratio $r$ of samples derived from random samplers is small and oracle samplers are dominant, the student robot performance was reasonably well. It can be also seen that the proposed DFRD scheme with the RRF feature space regularization does not deteriorate for large ratio $r\le 640$ \% for experiments considered here.

\bibliographystyle{IEEEtran} 
\bibliography{reference}

\end{document}